\documentclass[11pt,a4paper]{article}
\usepackage[hyperref]{acl2021}

\usepackage{times}
\usepackage{latexsym}
\usepackage{booktabs}
\usepackage{multirow}
\usepackage{amsmath, amsfonts, amssymb}
\usepackage{pifont}
\newcommand{\cmark}{\ding{51}}
\newcommand{\xmark}{\ding{55}}
\usepackage{graphicx}
\usepackage{multirow}
\usepackage{tabularx}
\usepackage{tabulary}
\usepackage{pdflscape}
\usepackage{rotating}
\usepackage{makecell}

\usepackage[T1]{fontenc}
\usepackage[utf8]{inputenc}

\usepackage{microtype}

\newcommand*\rot{\rotatebox{90}}

\aclfinalcopy

\title{Comparing Test Sets with Item Response Theory}

\author{Clara Vania$^{\spadesuit}$\thanks{~~Equal contribution.}~\hspace{0.1em}\thanks{\ \ Work done while at New York University.} ~~ 
Phu Mon Htut$^{\clubsuit}$\footnotemark[1] ~~ 
William Huang$^{\clubsuit}$\footnotemark[1] \\
\textbf{Dhara Mungra$^{\clubsuit}$ ~~ 
Richard Yuanzhe Pang$^{\clubsuit}$ ~~
Jason Phang$^{\clubsuit}$} \\
\textbf{Haokun Liu$^{\diamondsuit}$\footnotemark[2] ~~
Kyunghyun Cho$^{\clubsuit}$ ~~
Samuel R. Bowman$^{\clubsuit}$} \\
$^{\spadesuit}$Amazon ~~
$^{\clubsuit}$New York University \\
$^{\diamondsuit}$Allen Institute for AI \\
\texttt{vaniclar@amazon.co.uk, bowman@nyu.edu}
}

\date{}

\begin{document}
\maketitle
\begin{abstract}
Recent years have seen numerous NLP datasets introduced to evaluate the performance of fine-tuned models on natural language understanding tasks. Recent results from large pretrained models, though, show that many of these datasets are largely saturated and unlikely to be able to detect further progress.
What kind of datasets are still effective at discriminating among strong models, and what kind of datasets should we expect to be able to detect future improvements?
To measure this uniformly across datasets, we draw on Item Response Theory and evaluate 29 datasets using predictions from 18 pretrained Transformer models on individual test examples.
We find that Quoref, HellaSwag, and MC-TACO are best suited for distinguishing among state-of-the-art models, while SNLI, MNLI, and CommitmentBank seem to be saturated for current strong models. We also observe span selection task format, which is used for QA datasets like QAMR or SQuAD2.0, is effective in differentiating between strong and weak models.

\end{abstract}

\section{Introduction}
\label{sec:intro}

\begin{figure*}[t]
    \centering
    \includegraphics[width=0.95\linewidth]{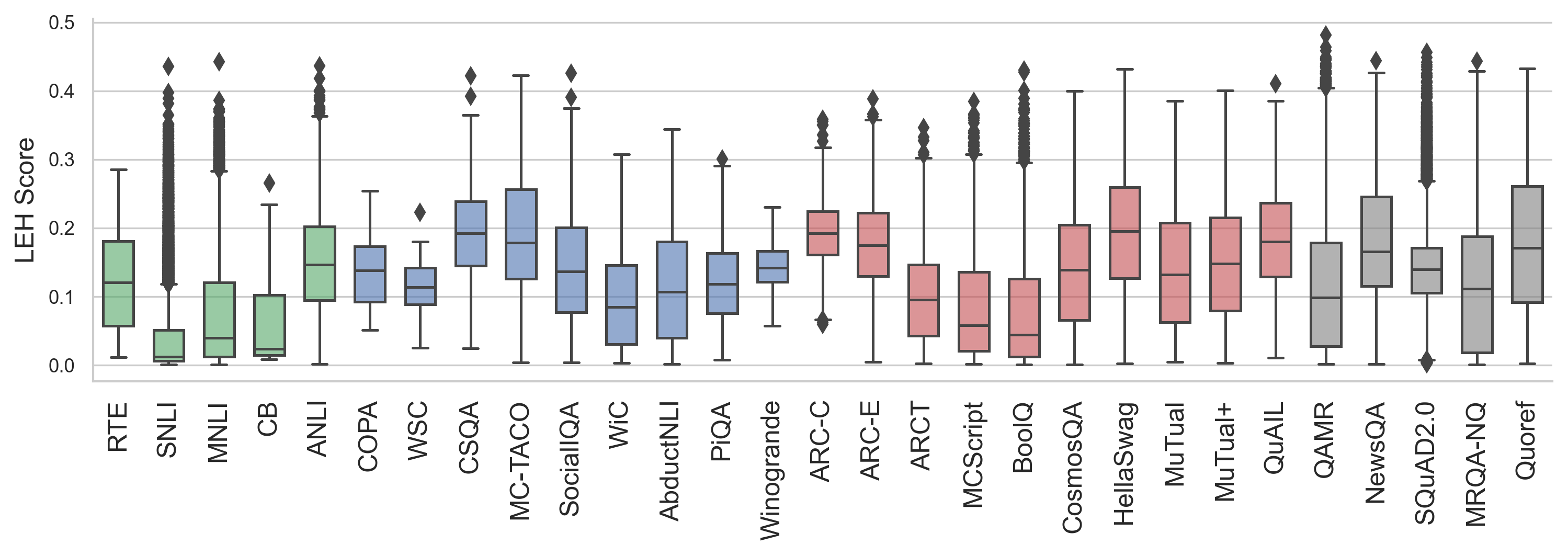}
    \caption{Distribution of test examples according to our proposed \textit{locally estimated headroom} (LEH) scores (\textsection~\ref{sec:item-params}), which measure the \textbf{local slope} of the Item Characteristic Curve (ICC) for an example at the ability level corresponding to the \textbf{best model}, and thus reflect the effectiveness of that single example at distinguishing between near-state-of-the-art models. Datasets are grouped by task format: classification (green), sentence-level multiple-choice (blue), paragraph-level multiple-choice (red), and span selection (grey). Within each format, the datasets are sorted by their release date. More details on the datasets are given in Table \ref{tab:tasks}.}
    \label{fig:irt-score-box}
\end{figure*}

Many datasets have been created to evaluate various aspects of natural language understanding (NLU) in English. 
These datasets are useful to measure progress; however, it is evident from various leaderboards \citep{wang-etal-2018-glue,SuperGLUE,rajpurkar-etal-2016-squad,zellers-etal-2018-swag} that many of them are no longer challenging or discriminative enough to differentiate strong models such as those based on Transformers \citep{vaswani-transformer-2017}.\footnote{For example, the recent DeBERTa model 
\citep{he2020deberta} achieves parity with human annotators on the SuperGLUE benchmark score: \url{https://super.gluebenchmark.com/leaderboard}.} 
Even if these benchmarks are sound tests of important (and potentially unsolved) tasks, their usefulness is limited if they cannot measure further progress. In this paper, we ask: Which datasets are best in distinguishing current and possible future strong models? 

We aim to compare datasets using a single metric that accounts for their  effectiveness in separating current stronger and weaker models. To that end, we use Item Response Theory \citep[IRT;][]{Baker1993ItemRT}, a statistical framework from psychometrics that is widely used for the evaluation of test items in educational assessment. IRT assumes that the probability that a model will correctly handle an example in a test set depends on the model's latent ability parameter and three example-specific parameters, typically measuring example difficulty (how strong does a model have to be to get it right), discrimination (how effective the example is for differentiating between similar models), and guessing (how likely a weak model is to get the example right for spurious reasons).

% \citet{lalor-etal-2016-building,lalor-etal-2018-understanding} use IRT to identify hard examples in natural language inference data based on human responses. However, since they focus on variation among human annotators (1,000 annotations per example), their analysis is limited to a small set of examples. In a follow-up study, \citet{lalor-etal-2019-learning} compare human versus model responses and find that both are positively correlated and demonstrate the use cases of IRT parameters in training set filtering. By using model responses, we can use IRT to evaluate a larger number of examples, as it is much easier to get a sufficient number of responses from models than from humans.

This paper presents a large-scale IRT analysis of existing English NLU datasets. Unlike previous work which focuses on example-level analysis \textit{within} individual datasets \citep{lalor-etal-2016-building,lalor-etal-2018-understanding}, here we analyze example characteristics from a larger perspective by comparing individual examples \textit{across} datasets. We evaluate test sets from 29 datasets in different formats---classification, multiple-choice QA, and span-selection QA.
As responses, we use model predictions from 18 Transformer-based models, including some limited-capacity models chosen to expose better the dataset's ability to discriminate weaker from stronger predictors. 
We then fit a single IRT model on these responses using a variational inference method.\footnote{
Our data and code can be found at \url{https://github.com/nyu-mll/nlu-test-sets}.} 

\newpage
We find: 
\begin{itemize}
    \item Quoref, HellaSwag, and MC-TACO contain the highest number of examples that can differentiate between near-state-of-the-art models, making them very likely to be effective at tracking near-future progress on the skills that they actually test (Figure~\ref{fig:irt-score-box}).
    \item SQuAD2.0, NewsQA, QuAIL, MC-TACO, and ARC-Challenge have the most difficult examples.
    \item Span-based QA is an effective task format for discriminating between strong and weak models.
    \item CosmosQA, MC-TACO, Winogrande, and ARC-Challenge consist mostly of hard examples, while for most datasets, the example difficulty levels are more widely distributed.
\end{itemize}

\section{Item Response Theory}
\label{sec:irt}

\begin{figure}[t]
    \centering
    \includegraphics[width=0.47\textwidth]{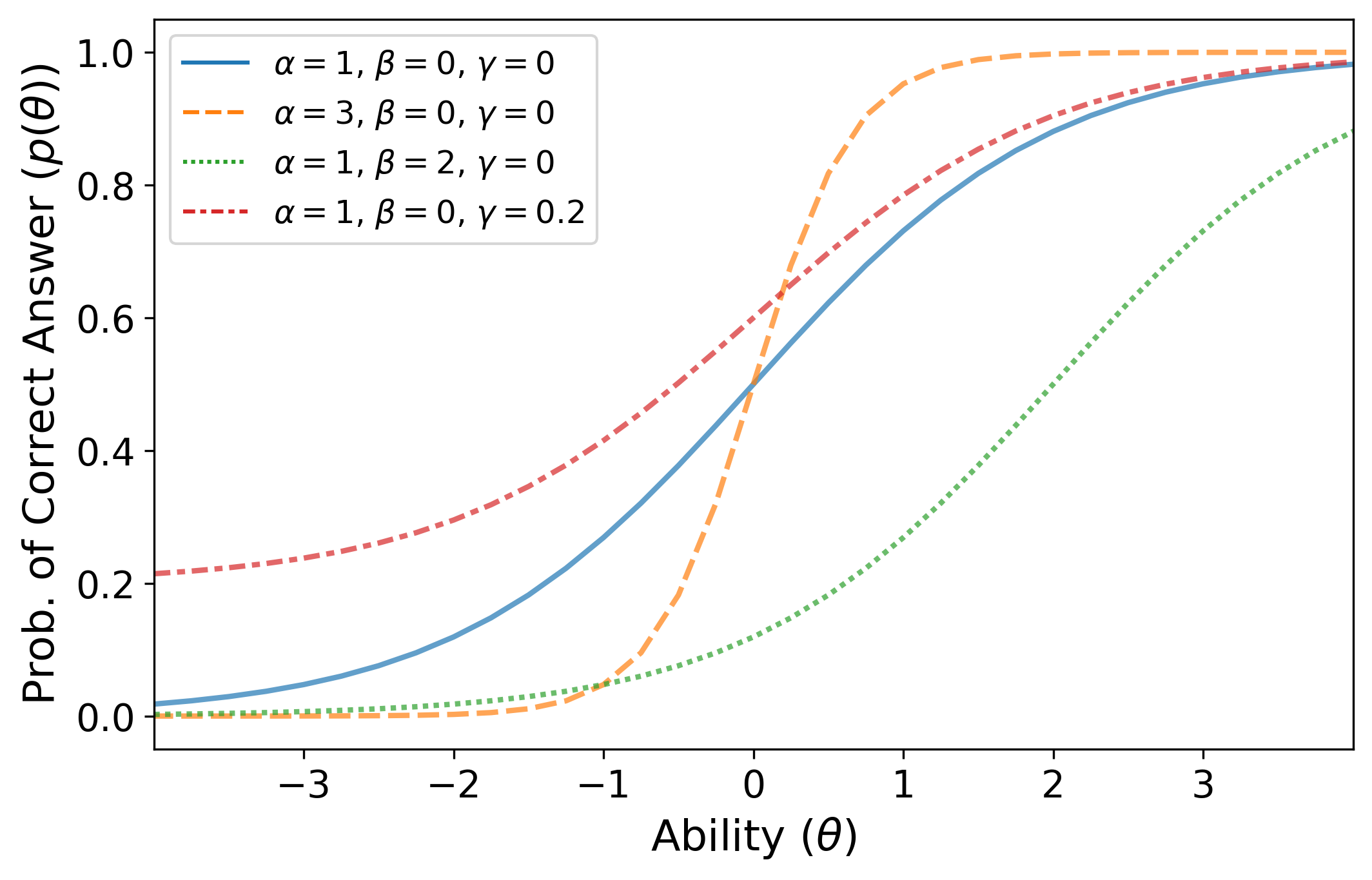}
    \caption{An example of item characteristic curves (ICCs) with different values for discrimination ($\alpha$), difficulty ($\beta$), and guessing ($\gamma$) parameters. $p(\theta)$ is the probability of a correct answer for a given $\theta$. $\theta$ measures a model's ability level (higher is better). $\alpha$ governs the steepness of the function, $\beta$ determines the $\theta$ value at which the curve is the steepest, while $\gamma$ defines the baseline likelihood that an arbitrarily weak model can guess correctly.}
    \label{fig:icc-example}
\end{figure}

\citet{Baker1993ItemRT} introduce Item Response Theory (IRT), a statistical framework to measure the probability of a responder (human or AI system) predicting a correct answer for a given item (test example).
The probability of a responder $i$ answering an item $j$ correctly is estimated as a function of the responder's latent ability $\theta_i$ and the item characteristics, referred to as the item characteristic curve (ICC). 

We use the 3-parameter (3PL) IRT model, where item behavior is governed by discrimination, difficulty, and guessing parameters. The discrimination parameter ($\alpha$) defines how effective an item is for distinguishing predictors along the ability axis. The difficulty parameter ($\beta$) defines a minimum level of ability at which we expect to see high responder performance. The guessing parameter ($\gamma$) defines the probability of correctly answering an item by random guessing. Figure~\ref{fig:icc-example} shows example ICCs with different parameter values. 

Formally, the probability of individual $i$ answering item $j$ correctly is modeled as:
\begin{align}
    p_{j}(\theta_i) = \gamma_j + \frac{1-\gamma_j}{1+e^{-\alpha_j(\theta_i - \beta_j)}}.
    \label{eq:ICC}
\end{align}

\subsection{IRT with Variational Inference}
\label{app:irt-setup}

We use variational inference to infer IRT parameters from model response patterns using Pyro \citep{ranganath2014BlackBoxVI,bingham2019pyro}. \citet{lalor-etal-2019-learning} found this method effective when fitting IRT models to responses on SNLI. 
Let $n$ be the number of items and let $m$ be the number of responders. The response patterns is ${\bf{Y}} \in \mathbb{R}^{n\times m}$, where the $i$-th row corresponds to responder $i$ and the $j$-th column corresponds to item $j$. We define $y_{ij}\in [0,1]$ as the response of model $i$ to item $j$, where $y_{ij}=1$ indicates a correct response and $y_{ij}=0$ indicates an incorrect response.
We approximate the joint probability of the parameters $\pi(\theta, \alpha, \beta, \gamma \mid \mathbf{Y})$ with a variational posterior:
\begin{align}
  q(\theta, \alpha, \beta, \gamma) = \prod_{i=1}^I \pi_i^\theta(\theta_i) \prod_{j=1}^J \pi_j^\alpha(\alpha_i) \pi_j^\beta(\beta_i) \pi_j^\gamma(\gamma_i)
\end{align}
where $\pi^\rho(\cdot)$ denotes the density for parameter $\rho$. 
For each parameter, we choose the following distributions:
\begin{align}
    \theta \sim \mathcal{N}(\mu_\theta, \sigma_\theta^2) \\
    \log \alpha \sim \mathcal{N}(\mu_\alpha, \sigma_\alpha^2) \\
    \beta \sim \mathcal{N}(\mu_\beta, \sigma_\beta^2) \\
    \mathrm{sigmoid}^{-1}(\gamma) \sim \mathcal{N}(\mu_\gamma, \sigma_\gamma^2)
\end{align}
We fit the posterior parameters by minimizing 
%the KL divergence between the approximate and true posterior distributions, which is equivalent to minimizing 
the evidence lower bound (ELBO). When calculating the ELBO, we weight the log-likelihoods of each item's parameter by the inverse of the item's dataset size to control for test set size.

Following \citet{lalor-etal-2019-learning}, we use a prior of $\mathcal{N}(0, 1)$ for $\theta$, $\beta$, and $\mathrm{sigmoid}^{-1}(\gamma)$. While \citet{lalor-etal-2019-learning} uses $\mathcal{N}(0, 10^3)$ for item parameter priors, we encountered degenerate runs and instead use $\mathcal{N}(0, 1)$. 
For $\log \alpha$, we use $\mathcal{N}(0, \sigma_\alpha^2)$ where we set $\sigma_\alpha$ by searching $[0.25, 0.5]$ by increments of $0.05$ and use the value yielding the highest ELBO after excluding degenerate runs.
We use a sigmoid transformation for $\gamma$ to constrain the guessing probability to $(0, 1)$.

\begin{table*}[ht]
\setlength{\tabcolsep}{2.5pt}
    \centering
    \small
    \begin{tabular}{llrrrclrr}
    \toprule
    & & \bf{$|$Train$|$} & \bf{$|$Dev$|$} & \bf{$|$Test$|$} & \bf{Cust.} & \bf{Metric} & \bf{RoBERTa} & \bf{Human} \\
    \midrule
    & RTE \citep[et seq.]{dagan2005pascal} & 2,490 & 138 & 139 & \cmark & Acc. & 87.6 & 93.6 \\ 
    & SNLI \citep{bowman-etal-2015-large} & 550,152 & 10,000 & 10,000 &   & Acc. & 92.7 & -- \\ 
    & MNLI \citep{williams-etal-2018-broad} & 392,702 & 9,823 & 9,824 & \cmark & Acc. & 89.7 & 92.0 \\ 
    \rot{\rlap{\textbf{\makecell[c]{Classifi-\\cation}}}} 
    & CommitmentBank \citep[CB;][]{de2019commitmentbank} & 250 & 28 & 28 & \cmark & Acc. & 90.5 & 95.8\\ 
    & ANLI \citep{nie-etal-2020-adversarial} & 1,105,719 & 3,200 & 3,200 & & Acc. & 50.8 & -- \\ 
    \midrule
    & COPA \citep{roemmele2011choice} & 400 & 50 & 50 & \cmark & Acc. & 86.0 & 100.0 \\ 
    & WSC \citep{levesque2012winograd} & 554 & 52 & 52 & \cmark & Acc. & 78.8 & 100.0\\ 
    & CommonsenseQA \citep[CSQA;][]{talmor-etal-2019-commonsenseqa}  & 9,741 & 610 & 611 & \cmark & Acc. & 74.6 & 88.9\\ 
    & MC-TACO \citep{zhou-etal-2019-going} & 3,026 & 757 & 9,442 & \cmark & EM & 55.9 & 75.8 \\ 
    & SocialIQA \citep{sap-etal-2019-social} & 33,410 & 977 & 977 & \cmark & Acc. & 79.9 & 88.1 \\ 
    & WiC \citep{pilehvar-camacho-collados-2019-wic}  & 5,428 & 319 & 319 & \cmark & Acc. & 71.5 & 80.0\\ 
    & Abductive NLI \citep[AbductNLI;][]{Bhagavatula2020Abductive}  & 169,654 & 766 & 766 & \cmark & Acc. & 85.0 & 92.9 \\
    \rot{\rlap{\textbf{\makecell[c]{Sentence-Level\\Multiple Choice}}}} 
    & PIQA \citep{bisk2020piqa} & 16,113 & 919 & 919 & \cmark & Acc. & 77.6 & 94.9 \\ 
    & WinoGrande \citep{sakaguchi2020winogrande} & 40,398 & 633 & 634 & \cmark & Acc. & 77.3 & 94.0 \\ 
    \midrule
    & ARC-Easy \citep{clark2018think} &  2,251 & 570 & 2,376 &   & Acc. & 62.5 & -- \\
    & ARC-Challenge \citep{clark2018think} &  1,119 & 299 & 1,172 &  & Acc. & 37.5 & -- \\ 
    & ARCT \citep{habernal-etal-2018-argument} &  1,211 & 317 & 445 &   & Acc. & 86.7 & 79.8 \\ 
    & MCScript \citep{ostermann-etal-2018-mcscript} &  14,191 & 2,020 & 3,610 &   & Acc. & 92.8 & 98.2 \\ 
    & BoolQ \citep{clark-etal-2019-boolq} & 9,427 & 1,635 & 1,635 & \cmark & Acc. & 85.7 & 89.0 \\
    & Cosmos QA \citep{huang-etal-2019-cosmos} & 25,262 & 1,492 & 1,493 & \cmark & Acc. & 79.4 & 94.0 \\ 
    & HellaSwag \citep{zellers-etal-2019-hellaswag} & 39,905 & 5,021 & 5,021 & \cmark & Acc. & 84.1 & 95.6 \\
    \rot{\rlap{\textbf{\makecell[c]{Paragraph-Level\\Multiple Choice}}}}  
    & MuTual \citep{cui-etal-2020-mutual} & 7,088 & 443 & 443 & \cmark & Acc. & 87.8 & 93.8 \\ 
    & MuTual+ \citep{cui-etal-2020-mutual} & 7,088 & 443 & 443 & \cmark & Acc. & 77.9 & 93.0\\ 
    & QuAIL \citep{rogers2020getting} & 10,246 & 2,164 & 556 &   & Acc. & 73.3 & -- \\ 
    \midrule
    & QAMR \citep{michael-etal-2018-crowdsourcing} & 50,615 & 18,908 & 18,770 &   & EM & 79.6 & -- \\
    & NewsQA \citep{trischler-etal-2017-newsqa}  & 76,568 & 4,343 & 4,293 &   & EM & 57.8 & 46.5 \\ 
    & SQuAD2.0 \citep{rajpurkar-etal-2018-know}  & 130,319 & 5,675 & 6,198 & \cmark & EM & 91.5 & 86.8 \\ 
    & MRQA-NQ \citep{kwiatkowski-etal-2019-natural}  & 104,071 & 6,418 & 6,418 & \cmark & EM & 69.9 & -- \\ 
    \rot{\rlap{\textbf{\hspace{1mm} \makecell[c]{Span \\ Selection}}}} 
    & Quoref \citep{dasigi-etal-2019-quoref}  & 19,399 & 1,209 & 1,209 & \cmark & EM & 78.7 & 93.0 \\ 
    \bottomrule
    \end{tabular}
    \caption{Datasets grouped by their task format and ordered by release year. \textbf{Cust.} denotes cases when we use our own custom split. \textbf{Metric}: evaluation metric used in this study. \textbf{RoBERTa}: model performance using RoBERTa$_{\rm{Large}}$. \textbf{Human}: human performance.%\textbf{N/A}: not available. 
    %For MNLI, $|$dev$|$ and $|$test$|$ combine the matched and the mismatched portions.
    }
    \label{tab:tasks}
\end{table*}   

\section{Experiments}

\begin{figure*}
    \centering
    \includegraphics[width=\linewidth]{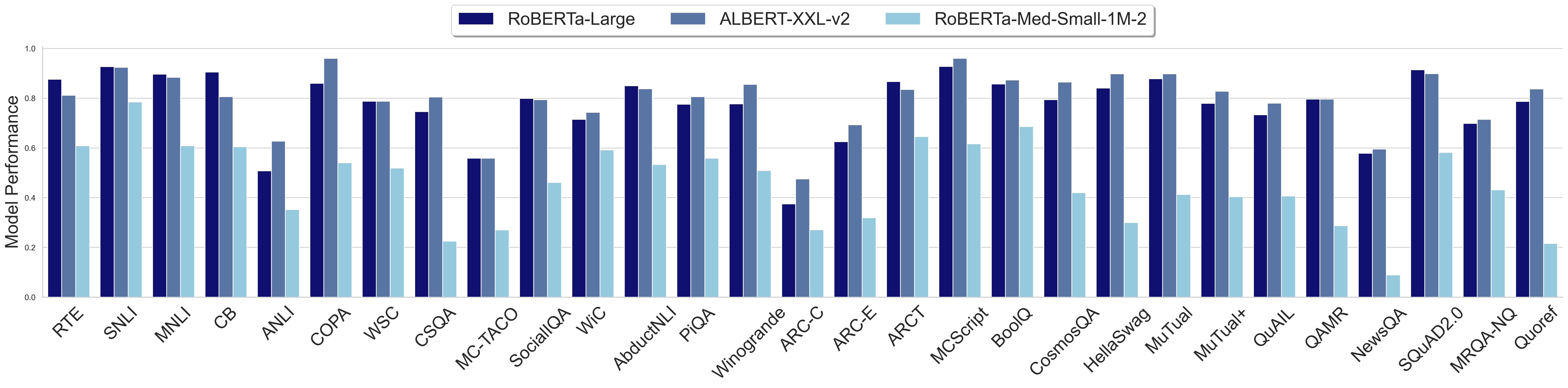}
    \caption{The best validation performance of ALBERT-XXL-v2, RoBERTa$_{\rm{Large}}$, and the smallest MiniBERTa (RoBERTa-Med-Small-1M-2) on each dataset. The full results table with performance of all models is reported in the Appendix (Table~\ref{tab:result_all_tasks})}
    \label{fig:exp-results}
\end{figure*}

\subsection{Datasets}

Our goal is to perform a fine-grained evaluation of English NLU datasets that appear to discriminate among widely used Transformer-based models. To that end, we choose datasets based on the following criteria:
\begin{itemize}
    \item They are plausibly unsolved, in that the best-reported model performance does not exceed estimated human performance (if available) by more than three metric points.
    \item They are relatively easy to use with current large pretrained models, and in particular, their inputs fit within a typical pretrained Transformer's  512-token limits. (This rules out tasks with full-document contexts or retrieval components.)
    \item They are evaluated at example-level, i.e., we focus our analysis on QA and other classification datasets, where each example corresponds to one item in the IRT. (This rules out structured prediction and sequence tagging tasks.)
    \item They have simple and reliable automatic metrics at the example level. (This rules out generation-based tasks.)
\end{itemize}
Table~\ref{tab:tasks} lists the datasets we evaluate. For MNLI, we combine the  matched and mismatched portions of the development and custom test sets for our analysis. For ANLI, we train models on SNLI, MNLI, and ANLI training examples. Similar to MNLI, we combine ANLI's three evaluation rounds of the development and the test sets for our analysis.

\paragraph{Custom Test Splits}
Some of our selected datasets do not have publicly available labeled test examples. For such cases, we create a new custom split by randomly sampling 50\% of the validation examples as a new test set and keeping the rest for validation (``Cust.'' column in Table~\ref{tab:tasks}). For Natural Questions, we use the MRQA 2019 version \citep{fisch-etal-2019-mrqa}, as the original version includes some examples with very long contexts.\footnote{
\url{https://github.com/mrqa/MRQA-Shared-Task-2019}
} 
For MC-TACO, the original dataset does not come with a training set. For our experiment, we use 80\% of the validation set as our training set and the rest as a our validation set while leaving the original test set untouched.

\subsection{Models}
\label{subsec:models}
 
We aim to understand how examples from different datasets contribute to the evaluations of models with near-state-of-the-art abilities, so we include several pretrained Transformer-based models to approximate this. However, using only high-performing models could result in a poor IRT model fit \citep{MARTINEZPLUMED201918} % with uninformative horizontal ICCs. 
To avoid this, we add both weaker models and under-trained versions of our original models. We use ALBERT-XXL-v2 \citep{Lan2020ALBERT}, RoBERTa$_{\rm{Large}}$ and RoBERTa$_{\rm{Base}}$ \citep{liu2019roberta}, BERT$_{\rm{Large}}$ and BERT$_{\rm{Base}}$ \citep{devlin-etal-2019-bert}, XLM-R \citep{conneau-etal-2020-unsupervised}, and 12 MiniBERTas \citep{Zhang20MiniBERTas}.
\footnote{The MiniBERTas are RoBERTa models pretrained on 1M, 10M, 100M, or 1B words of raw text, and varying slightly in model size. There are three pretrained models for each pretraining data quantity, which are pretrained using different near-optimal hyperparameter values. We use all three variants in producing responses for IRT.} 
For each of the 18 Transformer-based models, we evaluate five different checkpoints---at 1\%, 10\%, 25\%, and 50\% of the maximum steps of the maximum epochs (Section~\ref{subsec:setup}), as well as the best checkpoint on the validation set, which need not be one of the other four. This yields a total of 90 model predictions for each test example.

\subsection{Experimental Setup}
\label{subsec:setup}

\paragraph{Optimization} 

We perform a hyperparameter sweep on each dataset, varying the learning rate $ \in \{1e-5, 3e-5, 5e-6\}$. We tune the maximum epochs $ \in \{10, 40\}$ for small datasets ($<5$k training examples), and $\in \{3, 10\}$ for other datasets \citep{Zhang20Revisiting}. We use the \texttt{jiant} \citep{pruksachatkun-etal-2020-jiant} library which is based on PyTorch \citep{pytorch2019} and HuggingFace Transformers \citep{wolf-etal-2020-transformers}.

We only perform hyperparameter tuning with the RoBERTa$_{\rm{Large}}$ model and apply the best configuration to train all the other Transformer models. We use NVIDIA V100 Tensor Core GPUs for our experiments. On average, it takes approximately four hours to train RoBERTa on small datasets ($<3$k training examples), one day for medium-sized datasets ($<10$k), and four days for large datasets ($>10k$).

\section{Results and Analysis}

Figure~\ref{fig:exp-results} shows the performance of RoBERTa$_{\rm{Large}}$, ALBERT-XXL-v2, and one of the low performing MiniBERTas (RoBERTa-Med-Small-1M-2) on all validation sets. Unsurprisingly, ALBERT-XXL-v2 and RoBERTa$_{\rm{Large}}$ are the best-performing models, while the small MiniBERTa model achieves much lower performance. %For datasets with span selection format, we report the best exact match performance. For all other datasets, we report the respective standard evaluation metrics. 
Full results using all 18 models can be found in the Appendix (Table~\ref{tab:result_all_tasks}).

\subsection{IRT Analysis}

\subsubsection{Item Characteristics}
\label{sec:item-params}

\begin{figure*}[t]
    \centering
    \includegraphics[width=0.98\linewidth]{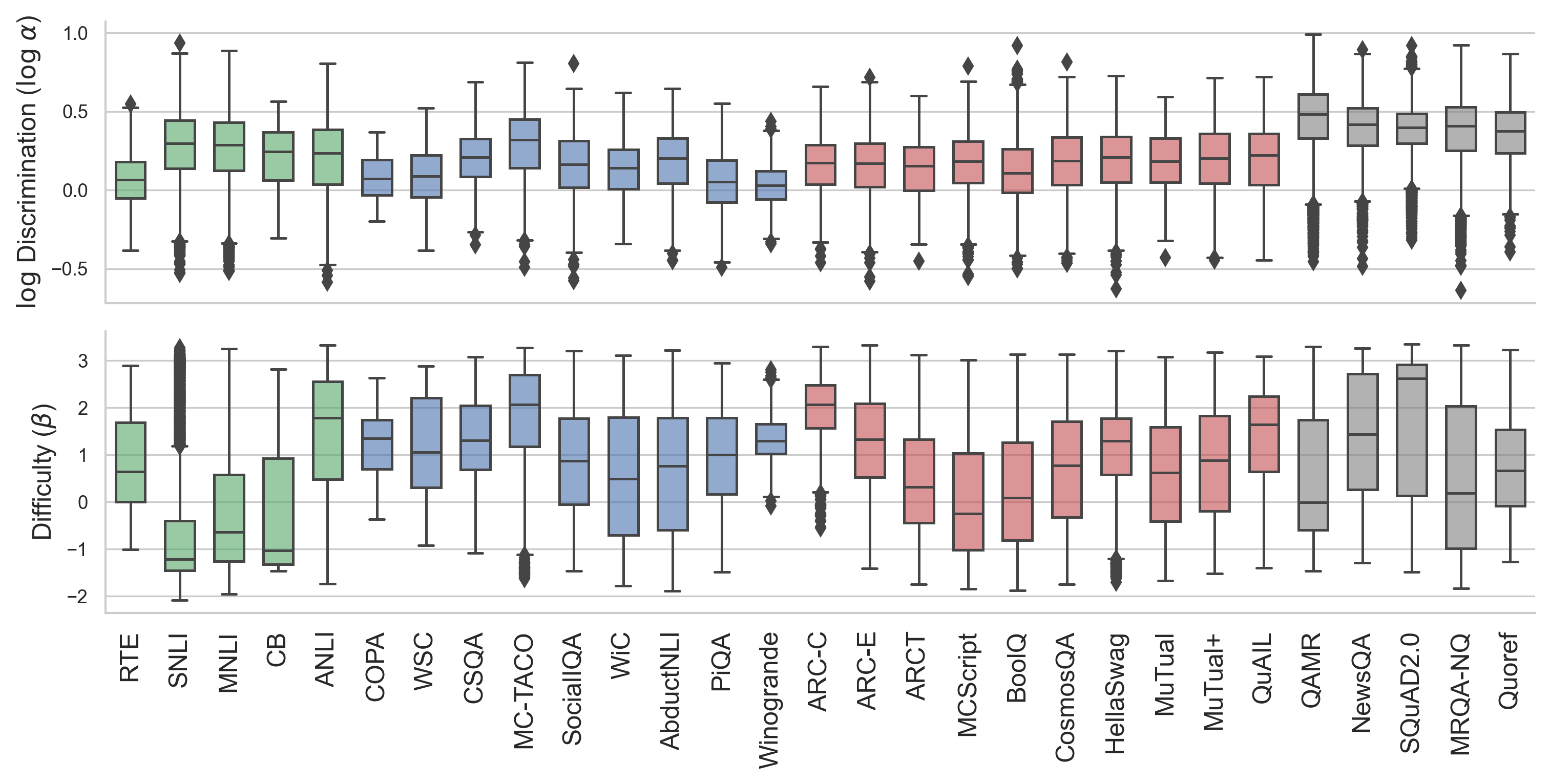}
    \caption{Distribution of test examples for each dataset based on the log discrimination ($\log \alpha$) parameter (top) and the difficulty ($\beta$) parameter (bottom).}
    \label{fig:disc-diff-box}
\end{figure*}

\begin{figure*}[t]
    \centering
    \includegraphics[width=0.95\linewidth]{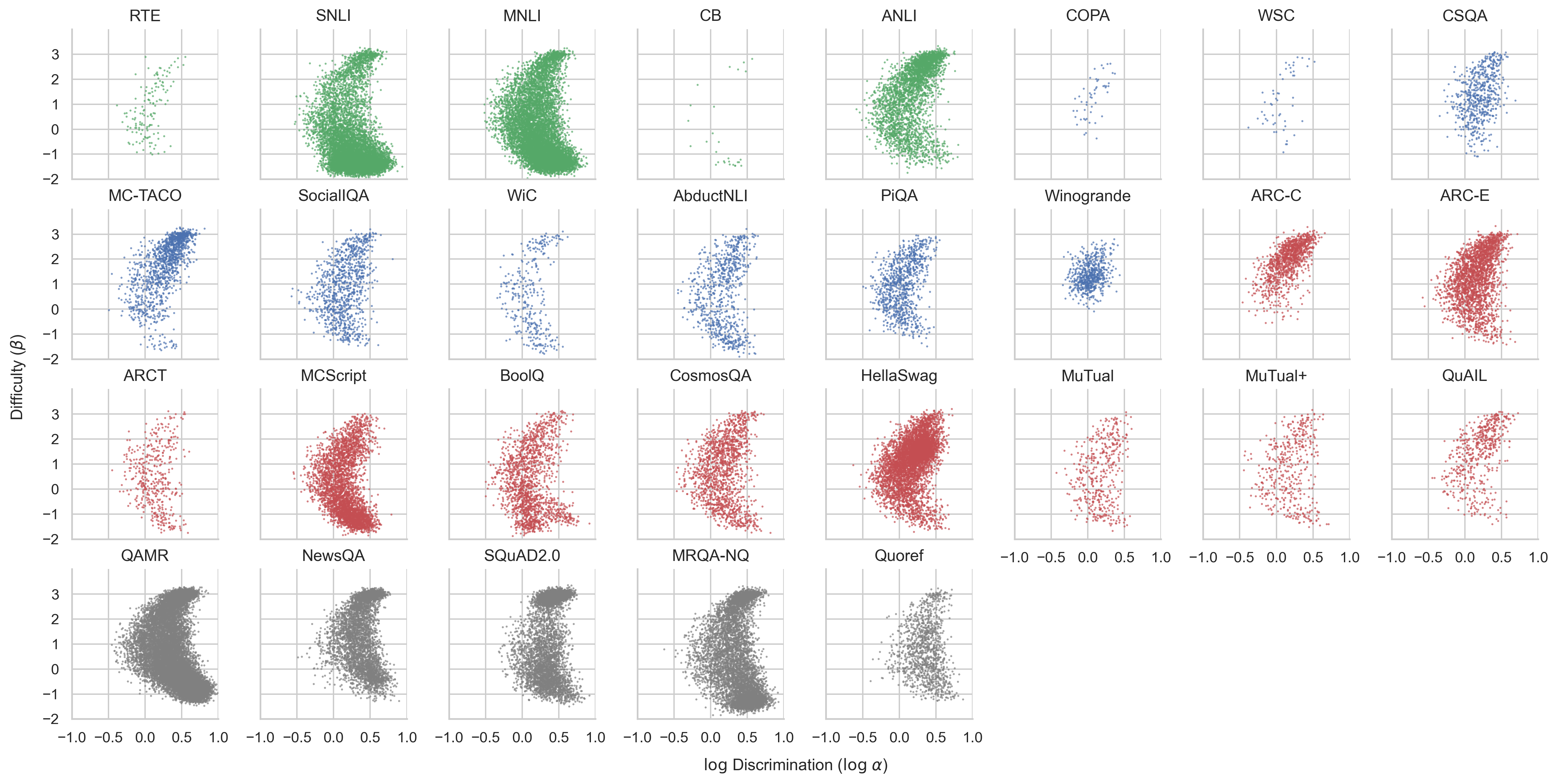}
    \caption{Distributions of log discrimination ($\log \alpha$) versus the difficulty ($\beta$) parameters for each dataset..}
    \label{fig:alpha-beta}
\end{figure*}

\paragraph{Metric}

As our primary metric, we introduce \textit{Locally Estimated Headroom} (LEH) score, which measures the ability of each test example to contribute to the evaluation of near-future progress. We calculate it as the derivative of the example's ICC (Figure~\ref{fig:icc-example}) with respect to the highest latent ability score, which corresponds to ALBERT-XXL-v2. 
A high LEH score indicates that the best-performing model is still far from the example's saturation points---the flat sections of ICC inferred by our model. There is enough space along the curve that the IRT model expects the example to be able to differentiate future state-of-the-art models. Typically, different near-state-of-the-art models both succeed and fail on this kind of example, while weaker models mostly fail. A high LEH score implies that there is still enough room for potentially stronger models to perform better on this dataset.

To validate the use of LEH scores for detecting near-future improvements, we compare two IRT models. The first is fitted using responses from all models, while the second is fitted based on responses from BERT and other weaker models (excluding  RoBERTa$_{\rm{Large}}$, RoBERTa$_{\rm{Base}}$, XLM-R, and ALBERT-XXL-v2). After that, we compute the correlation between the two sets of LEH scores, focusing on the 75\textsuperscript{th} percentile for each dataset. The Pearson correlation is 95.5\% with a median absolute difference of 0.007 and a standard deviation of 0.011. Out of the 29 datasets, only SQuAD2.0, CommensenseQA, MuTual, Quoref, and HellaSwag have more than 0.02 absolute difference in LEH scores. This strong correlation suggests that our ICCs fits are not overly sensitive to the exact characteristics of current state of the art models. 

\paragraph{Analysis by LEH Scores}

Figure~\ref{fig:irt-score-box} shows the distribution of test examples for each dataset based on their LEH scores. For our analysis, we focus on the 75$^{\rm{th}}$ percentile examples in each dataset as a rough proxy for how likely a dataset is to have a significant number of examples that are difficult or discriminative for near-future models. 

We observe that Quoref, HellaSwag, and MC-TACO have examples with the highest LEH scores, suggesting sufficient headroom for future state-of-the-art models with a higher ability to achieve better performance on these datasets. SNLI, CommitmentBank, and MNLI have relatively low LEH scores, indicating that performance on these datasets is largely saturated. Additionally, we also measure how the 75\textsuperscript{th} percentile LEH scores correlate with human-RoBERTa gap. Using 22 datasets that have human performance numbers (Table \ref{tab:tasks}), we find that the Pearson correlation between the two is weakly positive (0.21).

\paragraph{Analysis by Item Parameters}

Next, we analyze the distribution of test examples according to their discrimination and difficulty parameters (Figure~\ref{fig:disc-diff-box}). We observe that datasets with span selection format (QAMR, NewsQA, SQuAD, MRQA-NQ, and Quoref) have the highest discrimination scores than other datasets, highlighting span selection as an effective task format for discriminating among strong and weak models. 
However, this might be because this task format typically features a much larger space of possible model outputs than the other formats we consider. It does not necessarily mean that span selection is the most suitable to test models' ability to understand language. As the span-based format restricts answers to be text spans in the given passage, there are concerns that it rarely requires reasoning ability which often involves answers not mentioned in the passage, and thus not reflecting comprehension ability of humans \citep{lai-etal-2017-race,sugawara-etal-2018-makes}.

For the difficulty parameter, we do not observe a narrow task format that is superior to the others. However, we notice that the highest difficulty scores are obtained by QA datasets such as SQuAD2.0, NewsQA, QuAIL, ARC-Challenge, and MC-TACO. 
ANLI, which is created with adversarial model-in-the-loop crowdsourcing, also has of many hard examples.
Impressionistically, training set size and creation date do not seem to correlate with either example's difficulty or discrimination parameters.

Figure~\ref{fig:alpha-beta} shows the distribution of examples jointly according to their difficulty and log discrimination parameters. We notice a half-moon shape pattern in most datasets, which indicates that most of the discriminative examples are either very easy or very difficult. Referring to the ICC curve (Figure~\ref{fig:icc-example}), this indicates that there is high agreement among strong models or weak models, which corresponds to one of the saturation points in the ICC curve (upper or lower). The only dataset that does not have this pattern is Winogrande, which is difficult for all models. 

ARC-Challenge, QuAIL, HellaSwag, CommonsenseQA, and MC-TACO show clusters with high density on the top right regions, indicating a large number of examples with high discrimination and difficulty scores. Other datasets have more scattered distributions. SNLI, MNLI, and MCScript show higher density on the bottom right regions, while NewsQA, SQuAD2.0, and MRQA-NQ show higher density on both the top and bottom right regions. Further analysis of the guessing parameters can be found in Appendix~\ref{app:alpha-gamma}.

\subsection{Examples with Unanimous Responses}

When fitting ICC on examples that have only correct responses or only incorrect responses, the discrimination parameter is unconstrained. We find that these examples make up 4\% of our data. 13 of the 29 datasets contain at least one such example. Roughly 16\% of NewsQA examples are incorrectly answered by all models, while the remaining 12 datasets have less than 10\% of all correct or incorrect examples. To study the effect of examples with all correct or incorrect responses, we fit an IRT model on responses excluding such examples and compare against parameters from the full set of responses. We find that the Pearson correlation for the discrimination at the 75\textsuperscript{th} percentile is 97.2\%, with a median absolute difference of 0.016 and standard deviation of 0.015. MC-TACO, CommitmentBank, and WSC differ by more than 0.04. Further, we find that the Pearson correlation for the LEH score at the 75\textsuperscript{th} percentile is 98.9\%, with a median absolute difference of 0.006 and standard deviation of 0.005. RTE, WiC, WinoGrande, QAMR, NewsQA, MRQA-NQ, MC-TACO, and BoolQ differ by 0.01. Given these high correlations, we do not exclude these examples when reporting our main results.

\subsection{Analysis by Task Group}

Next, we analyze each task-type group in more detail, focusing on the example's scores around the 75$^{\rm{th}}$ percentile.

\paragraph{Classification} 

We observe that all datasets have moderate discrimination scores. Most ANLI examples have relatively high difficulty scores, while SNLI, MNLI, and CommitmentBank have the lowest difficulty scores.

\paragraph{Sentence-Level Multiple Choice}

All of the datasets in this group have relatively low discrimination scores compared to span selection datasets. Figure~\ref{fig:alpha-beta} shows that MC-TACO, Winogrande, and CommonsenseQA all have a higher density of difficult examples, while for other datasets the distribution is more spread.

\paragraph{Paragraph-Level Multiple Choice}

QuAIL and ARC-Challenge examples have high difficulty but moderate discrimination scores. As seen in  Figure~\ref{fig:alpha-beta}, these datasets have a higher density in the top right regions, showing a large proportion of difficult examples. ARCT shows moderate difficulty despite its known artifacts \citep{niven-kao-2019-probing}, indicating that it can still be challenging for models. Compared to other datasets, BoolQ has the highest number of easy examples. However, as it is a binary classification task, the random baseline performance is already high.

To investigate this, we calculate the number of examples in each test set that have $\gamma$ parameter below $0.5$. In general, we find that 88\% of the test examples have $\gamma < 0.5$, implying that most of the examples contributed to the inferences of $\alpha$, $\beta$, and $\theta$. BoolQ was the only exception in which approximately 56\% of examples were assigned $\gamma > 0.5$. After filtering out these guessable examples in BoolQ, we find that its test examples have slightly higher discrimination scores with little change in difficulty scores.

\paragraph{Span Selection} 

We observe that span selection datasets are the most discriminative. However, in terms of difficulty, only SQuAD2.0 and NewsQA are among the top five.

\begin{table*}[t]
    \centering
    \small
    \newcolumntype{L}{>{\arraybackslash}m{.75\textwidth}}
    \newcolumntype{K}{>{\arraybackslash}m{.05\textwidth}}
    \begin{tabular}{KLc}
    \toprule
    \textbf{Name} &  \textbf{Example} & \textbf{Difficulty ($\beta$)} \\
    \midrule
    MNLI & \textit{Premise}:  And, you know, with this, you know, it wasn't many opportunities for kids to be special, because kids weren't, you know, you were pushed out of adult conversation, and just really pushed to the side. & 3.27 \\
    \addlinespace[.10cm]
    & \textit{Hypothesis}:  Children were pushed out of adult conversation, and really just pushed to the side in general. & \\
    \addlinespace[.10cm]
    & \textit{Label}:  entailment & \\ 
    \midrule
    MNLI & \textit{Premise}:  Look, it's your skin, but you're going to be in trouble if you don't get busy. & -1.87 \\
    \addlinespace[.10cm]
    & \textit{Hypothesis}:  The boss will fire you if he sees you slacking off. & \\
    \addlinespace[.15cm] 
    & \textit{Label}:  neutral & \\
    \midrule
    \midrule
    MC-TACO & The Beatles are giving a press conference about their new film, Magical Mystery Tour .What time of day was the press conference? & 2.86 \\
    \addlinespace[.10cm]
    & (1) 4:00 PM \cmark \quad 
    (2) 12:00 PM \cmark \quad 
    (3) 3 p.m \cmark \quad 
    (4) 6:00 AM \xmark  
    & \\
    \midrule
    MC-TACO & Because then they feel like they are forced to stay in that situation."On average, how often do they feel stuck in the situation? & -1.67 \\
    \addlinespace[.10cm]
    & (1) 54 months \xmark \quad 
    (2) 6 centuries \xmark \quad 
    (3) once every 6 years \xmark  & \\
    \addlinespace[.10cm]
    & (4) every few seconds \xmark \quad 
    (5) once every 2 seconds \xmark \quad 
    (6) once every 18 years \xmark 
    & \\
    \bottomrule
    \end{tabular}
    \caption{Hardest and easiest examples along with their estimated difficulty score for MNLI and MC-TACO.}
    \label{tab:diff-examples}
\end{table*}

\subsubsection{Analysis on Model Ability}
\label{subsec:theta-analysis}

For a sanity check, we further analyze how each model scores according to our fitted IRT parameters. We observe a positive correlation between ability and average model accuracy (Appendix \ref{app:theta-analysis}). Generally, within a model, the best validation checkpoint obtains the highest average model accuracy and/or ability score. Across models, ALBERT-XXL-v2 performs typically best.

\subsection{Qualitative Analysis}

To better understand what kinds of examples are difficult or discriminating, we analyze the 20 examples with the lowest and highest scores for the discrimination and the difficulty parameters from five datasets: SQuAD2.0, MC-TACO, QuAIL, MNLI, and BoolQ. The first three are datasets with high discrimination and/or difficulty scores. MNLI and BoolQ have moderate discrimination and difficulty scores and low label entropy (three-class classification for MNLI and binary choice for BoolQ).

We observe that the 20 most difficult BoolQ examples are labeled \textit{False} (the minority class), while 19 of the 20 easiest examples are labeled \textit{True}. For MNLI, we find that the 20 easiest MNLI examples are labeled \textit{neutral} while the 20 hardest examples are a mixture of \textit{entailment} and \textit{contradiction}. 

In MC-TACO, each example contains a varying number of answer choices. For each choice, a model needs to predict whether the answer is \textit{True} or \textit{False}. We find that all answer choices in top 20 easiest examples are labeled \textit{False} (the majority class), whereas for difficult examples the answer choices are either all \textit{True} or a mix of \textit{True} and \textit{False} (Table~\ref{tab:diff-examples}). For SQuAD2.0 and QuAIL, we analyze the context length, the answerability of a question, and the lexical overlap between context and questions. However, we do not find any clear evidence that any of them might indicate the difficulty level of test examples. 

For BoolQ, we observe that the 20 most discriminating examples are all labeled \textit{False} while 13 of the 20 least discriminating examples are labeled \textit{True}. 
% We do not observe any clear patterns that can indicate discriminating examples in all the other four datasets.
Table~\ref{tab:diff-examples} shows the hardest and the easiest examples of MNLI and MC-TACO.
%, while Table~\ref{tab:disc-examples} in the Appendix shows the most and the least discriminating examples of MNLI and MC-TACO.

\section{Related Work}

Prior work on using IRT to evaluate NLP systems mostly relies on human responses. \citet{hopkins-may-2013-models} use IRT to estimate the relative ability of a set of machine translation systems using responses from pairwise comparison of system outputs by human judges. \citet{otani-etal-2016-irt} extend this work by including a baseline translation to the pairwise comparison. \citet{lalor-etal-2016-building,lalor-etal-2018-understanding} use IRT to identify hard examples in natural language inference data based on human responses. In a follow-up study, \citet{lalor-etal-2019-learning} compare human versus model responses and find that both are positively correlated and demonstrate the use cases of IRT parameters in training set filtering. \citet{sedoc-ungar-2020-item} use IRT to evaluate chatbot systems. 

The work by \citet{MARTINEZPLUMED201918} is the first to study the idea of using model responses (as opposed to human responses) for IRT in machine learning research.
For NLU, \citet{lalor-yu-2020-dynamic} use model responses to estimate difficulty parameters of several GLUE datasets for dynamic data selection in curriculum learning. In concurrent work, \citet{pedro-etal-2021-evaluation} study how IRT can be used for more nuanced leaderboard evaluations. Their experiments demonstrate that IRT can produce a more reliable ranking of models than the traditional metrics. They also show that IRT is not only useful for better understanding of individual examples in the dataset and task, but also effective in identifying annotation errors.

For other dataset evaluations, in addition to providing a benchmark, the SuperGLUE paper also compares a set of candidate datasets using a fixed pool of machine learning models and human annotators \citep{nangia-bowman-2019-human}.
\citet{wang-etal-2019-tell} investigate pretraining tasks and paradigms for effective transfer learning methods. \citet{pruksachatkun-etal-2020-intermediate} study when and why intermediate-task training is useful for a given target task. \citet{vu-etal-2020-exploring} introduce task embeddings to predict the most beneficial source task for a given target task. \citet{schlegel-etal-2020-framework} propose an evaluation framework for machine reading comprehension (MRC) datasets and reveal some concerns regarding factual correctness and the presence of linguistic cues in existing MRC gold datasets.

\section{Conclusion}

Given the large number of NLU datasets introduced in recent years, what kinds of datasets are effective to measure near-future progress? Our analysis on 29 test sets using IRT gives us reason to believe that, among the datasets we evaluate, Quoref, HellaSwag, and MC-TACO are best able to discriminate among current (and likely future) strong models. Meanwhile, SNLI, MNLI, and CommitmentBank seem to be saturated and ineffective for measuring future progress.

Our analysis of examples' difficulty and discrimination parameters shows that datasets with many hard examples do not always contain examples that can discriminate between strong and weak models. We find that QA datasets are more difficult than other datasets. We also find span selection as the most effective task format for discriminating between strong and weak models. 

According to our LEH score, datasets that seem to be solved are unlikely to see improvements with future pretrained models. Therefore, the skills they intend to test are either largely solved, to the extent that they are solvable, or not well isolated (e.g., due to data artifacts). Focusing on the skills for which these solved test sets are originally designed to evaluate would most likely require a new dataset that better isolates the reasoning ability of interest.

On the other hand, datasets that perform well according to our LEH metric show the best signs of being amenable to future hill-climbing. This \textit{does not} entail that we should focus future research on these benchmarks, since we do not evaluate whether they test the skills they mean to test, or whether these skills are important for scientific or practical progress on natural language understanding. Finally, we argue that this evaluation should be done periodically, as datasets and models improve over time.

For future work, one can study multi-dimensional variables for both model ability and item parameters, which could reveal a factorization of datasets by skills. Other potential directions include expanding our analysis to a broader range of tasks and analyzing the relationship between the estimated IRT parameters and the human-model gap.

\section*{Acknowledgments}

We thank John Lalor, João Sedoc, Nikita Nangia, Sebastian Schuster, Iacer Calixto, and the anonymous reviewers for feedback. 
This work has benefited from financial support to SB by Eric and Wendy Schmidt (made by recommendation of the Schmidt Futures program), Samsung Research (under the project \textit{Improving Deep Learning using Latent Structure}), Apple, and Intuit, and from in-kind support by the NYU High-Performance Computing Center and by NVIDIA Corporation (with the donation of a Titan V GPU). This material is based upon work supported by the National Science Foundation under Grant No.\ 1922658. Any opinions, findings, and conclusions or recommendations expressed in this material are those of the author(s) and do not necessarily reflect the views of the National Science Foundation. 

% \clearpage

\section*{Ethical Considerations}

We present an objective approach for comparing the difficulty of test sets examples across datasets and demonstrate it on a large set of established datasets. We expect this to contribute to the development of more challenging benchmarks for NLP datasets and potentially to develop more challenging models. One concern worth noting is that most of the evaluation datasets we study are crowdsourced or drawn from naturally occurring data. Thus, they likely demonstrate harmful stereotypes to some degree or even score models more highly for demonstrating them. In general, models that perform well on these datasets should not be deployed directly without additional measures to measure and eliminate any harms that stereotypes like these could cause in the target application settings.

\bibliography{anthology,acl2021}
\bibliographystyle{acl_natbib}

\appendix

\clearpage

\section{Discrimination vs. Guessing}
\label{app:alpha-gamma}

\begin{figure*}[t]
    \centering
    \includegraphics[width=\linewidth]{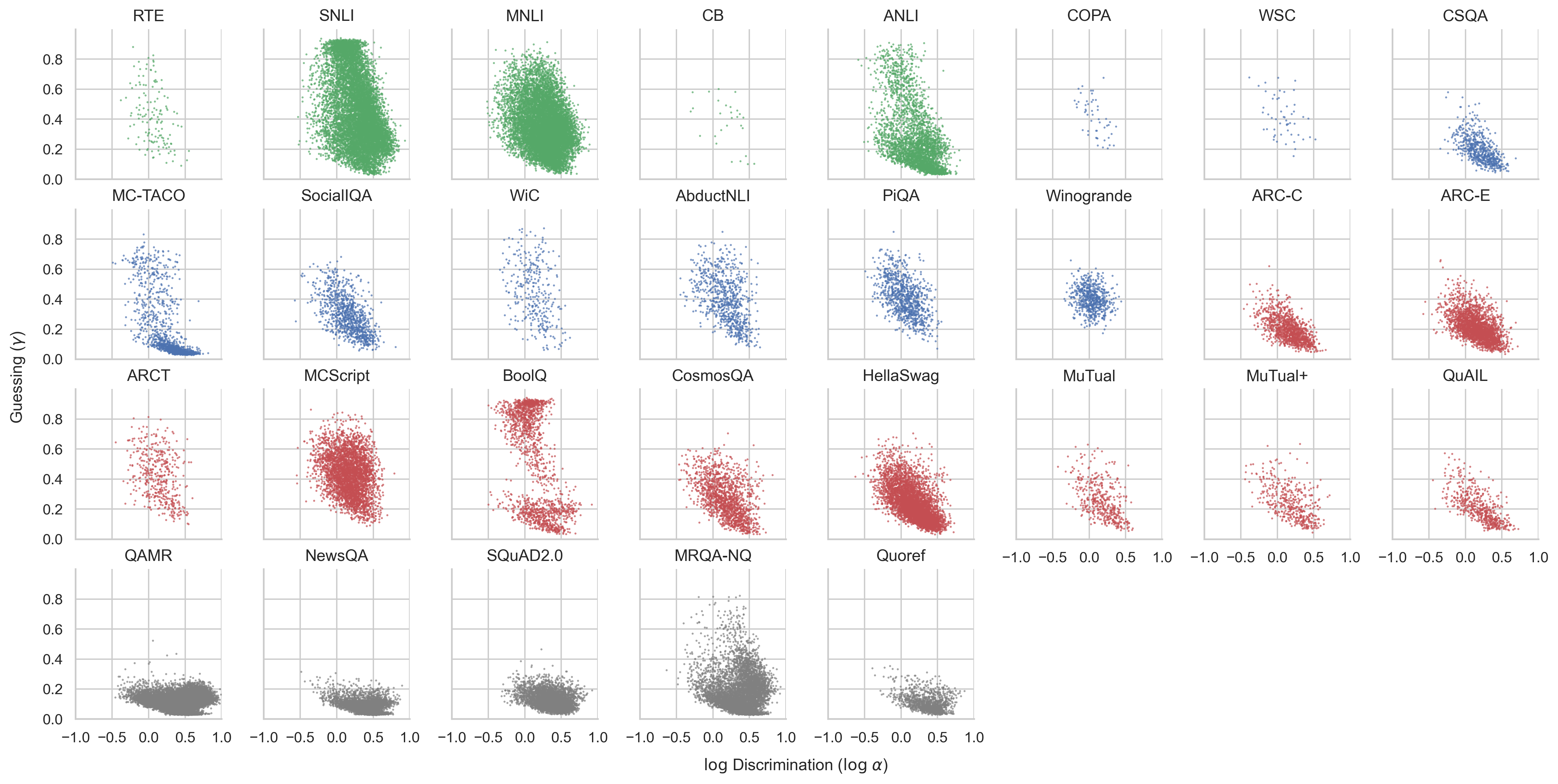}
    \caption{Plots of the log discrimination ($\log \alpha$) versus the guessing ($\gamma$) parameters for each dataset.}
    \label{fig:alpha-gamma}
\end{figure*}

In addition to the analysis of discrimination versus difficulty parameters, we also look at the distribution of the guessing ($\gamma$) parameters. From Figure~\ref{fig:alpha-gamma}, we observe that all QA datasets with span selection format generally have low guessing parameters, meaning that they are difficult to predict correctly by random guessing. This makes sense as span selection has higher label entropy than classification or multiple-choice task. We find that several datasets have examples with varying guessing parameters: For SNLI we see a high density of examples that can be predicted easily by random guessing while for MNLI, HellaSwag, and MCScript, there are more examples with low guessing parameters.

\begin{figure*}[t]
    \centering
    \includegraphics[width=0.97\linewidth]{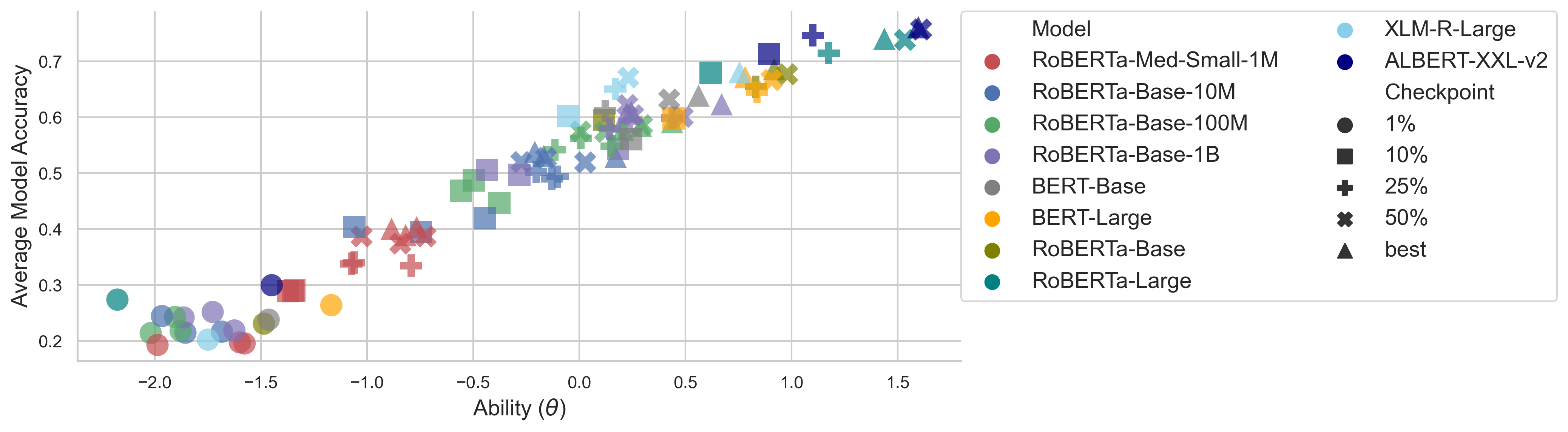}
    \caption{Average model accuracy over all datasets vs. ability ($\theta$). The three different hyperparameter configurations of each MiniBERTa are represented by a single color for ease of readability. Best viewed in color.}
    \label{fig:theta-acc}
\end{figure*}

\section{Additional Analysis on Model Ability}
\label{app:theta-analysis}

Figure~\ref{fig:theta-acc} plots model abilities $\theta$ against their average accuracy over all test examples, where each point represents a model checkpoint (Section~\ref{subsec:models}). We use different colors for different models (e.g., dark blue for ALBERT-XXL-v2), and different shapes to mark different checkpoints. 
% As expected, we observe a positive correlation between ability and average model accuracy.

% Unsurprisingly, models trained to 1\% of their tuned step limits have the lowest ability. 
% Within a model, generally the checkpoint selected based on validation-set early stopping obtains the highest average model accuracy and/or ability score. Across models, we find ALBERT-XXL-v2 outperforms other models.
Since we only perform tuning on RoBERTa$_{\rm{Large}}$, some of these models might have worse performance than if they were individually tuned.

\section{Task Descriptions}
 
In this section, we provide a short description for each dataset.

\paragraph{RTE} The series of Recognizing Textual Entailment datasets \citep{dagan2005pascal, haim2006second, giampiccolo-etal-2007-third, bentivogli2009fifth} correspond to a two-class textual entailment classification task. Given a premise sentence and a hypothesis sentence, the task is to decide whether the premise entails the hypothesis.

\paragraph{SNLI} The Stanford Natural Language Inference corpus \citep{bowman-etal-2015-large} is a textual entailment dataset, formulated as a three-class classification task. Given a premise sentence and a hypothesis sentence, the task is to determine if the premise entails the hypothesis, contradicts it, or neither. The SNLI dataset is created using premises taken from image captions.

\paragraph{MNLI} The Multi-Genre Natural Language Inference corpus \citep{williams-etal-2018-broad} is also a textual entailment dataset, similar to that of SNLI. The MNLI dataset is built to cover a broad range of genres, including written and spoken text. Half of its test set is created from text that is out of domain relative to the training set.

\paragraph{CommitmentBank} CommitmentBank \citep{de2019commitmentbank} is a dataset formulated as a three-class textual entailment classification task. Given a piece of text and an embedded clause, models must decide whether the embedded clause is entailed by the text.

\paragraph{ARCT} The Argument Reasoning Comprehension Task \citep{habernal-etal-2018-argument} is a multiple-choice question answering dataset. Given an argument, a claim, and a premise, the task is to select the correct implicit warrant (which explains why the premise implies the claim) from two choices.

\paragraph{ARC-Easy} ARC \citep{clark2018think} is a multiple-choice QA dataset composed of real multiple-choice science questions in grade schools. ARC-Easy is composed of the easier questions that do not satisfy the criteria used to built ARC-Challenge (described below).

\paragraph{ARC-Challenge} ARC-Challenge \citep{clark2018think} is the subset of ARC that contains questions that are incorrectly answered by both a retrieval-based algorithm and a word co-occurrence algorithm.

\paragraph{MCScript} The MCScript \citep{ostermann-etal-2018-mcscript} is a QA dataset with multiple-choice format. The dataset tests models' commonsense knowledge, in particular, script knowledge which corresponds to the sequence of actions people do in a particular situation. 

\paragraph{Cosmos QA} Cosmos QA  \citep{huang-etal-2019-cosmos} is a multiple-choice reading comprehension dataset, and it is intended to require extensive abstractive commonsense reasoning. Unlike CommonsenseQA, Cosmos QA requires comprehension over an auxiliary article, instead of simply responding to a free-standing question.

\paragraph{HellaSwag} HellaSwag \citep{zellers-etal-2019-hellaswag} is a commonsense reasoning multiple-choice dataset. It is built using adversarial filtering with BERT. Given a story, the task is to select the most plausible continuation.

\paragraph{BoolQ} BoolQ \citep{clark-etal-2019-boolq} is a boolean (yes/no) reading comprehension QA dataset built using the same pipeline used to produce the (non-boolean) Natural Questions \citep{kwiatkowski-etal-2019-natural}.

\paragraph{MuTual} MuTual \citep{cui-etal-2020-mutual} is a multiple-choice QA dataset for multi-turn dialogue reasoning. The dataset is created from Chinese students' English listening comprehension exams, and it is intended to require a variety of commonsense reasoning skills.

\paragraph{MuTual-Plus} MuTual-Plus  \citep{cui-etal-2020-mutual} is a variant of MuTual, in which one of the choices in each set of answers is replaced by a safe response (i.e., ``could you repeat that''). If all other choices are incorrect, then the model is supposed to select the safe response. This variant of MuTual is built so that we can evaluate if the model can select the safe response when all other options are incorrect.

\paragraph{QuAIL} QuAIL  \citep{rogers2020getting} is a reading comprehension dataset formulated as a multiple choice task. One feature of QuAIL is that it combines ``commonsense, text-based, and unanswerable questions.'' It is also designed such that it has a balanced distribution of genres and reasoning types.

\paragraph{COPA} Choice of Plausible Alternatives \citep{roemmele2011choice} is a dataset for sentence-level multiple-choice task. Given a premise and a question that asks for the cause or effect of the premise, the task is to choose the most plausible hypothesis from two options.

\paragraph{WSC} The Winograd Schema Challenge \citep{levesque2012winograd} is a sentence-level multiple-choice commonsense reasoning dataset. Given a piece of text, a pronoun, and a list of possible noun phrases, the model must choose the correct referent to the pronoun. The dataset is designed such that world knowledge is required to make the correct choices. We use the SuperGLUE \citep{SuperGLUE} version of the dataset.

\paragraph{CommonsenseQA} CommonsenseQA \citep{talmor-etal-2019-commonsenseqa} is a multiple-choice QA dataset which is designed to test a range of commonsense knowledge.

\paragraph{SocialIQA} SocialIQA \citep{sap-etal-2019-social} is a dataset that is specifically designed to test a models' capabilities related to emotional and social intelligence in everyday situations.

\paragraph{MC-TACO} MC-TACO \citep{zhou-etal-2019-going} is a multiple-choice QA dataset that is designed to test temporal commonsense reasoning, in particular: duration, temporal ordering, typical time, frequency, and stationarity. Each question consists of a varying number of choices, and for each answer choice, a model needs to predict whether the answer is correct or incorrect.

\paragraph{WiC} The Word-in-Context  \citep{pilehvar-camacho-collados-2019-wic} dataset which is designed to test the word sense disambiguation skill of a model. Given two pieces of text (a phrase or a sentence) with a polysemous word in both, a model needs to predict whether the two words are used in the same sense.

\paragraph{PIQA} The Physical Interaction Question Answering dataset \citep{bisk2020piqa} is a multiple-choice QA dataset that is designed to test the physical commonsense reasoning skill. Given a physical task expressed in text, a model needs to select the most sensible solution.

\paragraph{WinoGrande} The WinoGrande dataset \citep{sakaguchi2020winogrande} is built through a crowdsourcing procedure that incorporates adversarial filtering. Given a sentence with a blank (where the blank corresponds to a noun phrase), the task is to select the correct filler. The dataset is designed to test the commonsense reasoning skill. 

\paragraph{Abductive NLI} The Abductive Natural Language Inference dataset \citep{Bhagavatula2020Abductive} is a multiple-choice dataset. Given a premise, the task is to select the most likely explanation from the given hypotheses.

\paragraph{QAMR} The Question-Answer Meaning Representations \citep{michael-etal-2018-crowdsourcing} is a QA dataset where the question-answer pairs are created from sentences' predicate-argument relationships.

\paragraph{NewsQA} NewsQA \citep{trischler-etal-2017-newsqa} is a QA dataset formulated as span selection task. The dataset is built by crowdworkers using passages taken from CNN news articles. 

% \paragraph{MCTest \citep{richardson-etal-2013-mctest}} MCTest is a QA dataset formulated as a multiple choice task, which corresponds to open domain machine comprehension. The task requires models to answer questions about fictional stories. 

\paragraph{SQuAD2.0} SQuAD2.0 \citep{rajpurkar-etal-2018-know} is a QA dataset that combines the span-selection reading-comprehension questions in SQuAD 1.1 \citep{rajpurkar-etal-2016-squad} with over 50,000 unanswerable questions. The unanswerable questions were written by crowdworkers to look like the answerable ones. A model must either select an answer span or decline to answer. 

\paragraph{Quoref} Quoref \citep{dasigi-etal-2019-quoref} is a QA dataset that is designed to test coreferential reasoning ability. The dataset is formulated as a span selection QA task.

\paragraph{MRQA Natural Questions} The Natural Questions dataset \citep{kwiatkowski-etal-2019-natural} is a dataset designed to test a model's ability in reading comprehension. The questions are taken from real-word queries, while the context passages are taken from Wikipedia articles. We use the MRQA version of it which contains a preprocessed version of a subset of questions in Natural Questions.

\paragraph{ANLI} The Adversarial Natural Language Inference dataset \citep{nie-etal-2020-adversarial} is a textual entailment dataset built using an iterative human-and-model-in-the-loop procedure in order to find hard examples.

\clearpage

\clearpage
% \section{Validation Set Performance}
% \label{app:val-perf-table}

\begin{sidewaystable*}[t]
\small
\centering
\resizebox{0.9\linewidth}{!}{
    \begin{tabular}{lrrrrrrrrrrrrrrrrrrr}
    \toprule
    Dataset & Best & ALBERT & RL & RB & XLM-R & BL & BB &  RB- 100M-1  &   RB-100M-2   & RB-100M-3 & RB-10M-1    &  RB-10M-2   &  RB-10M-3 & RB-1B-1 & RB-1B-2  &   RB-1B-3 & RMS-1M-1   & RMS-1M-2  &   RMS-1M-3   \\
    \midrule
    RTE & 86.6 & 81.2 & 87.6 & 80.4 & 57.2 & 81.9 & 73.9 & 60.9 & 66.7 & 66.7 & 58.7 & 59.4 & 57.2 & 67.4 & 64.5 & 71.7 & 60.9 & 60.9 & 58.0 \\ 
    SNLI & 92.6 & 92.4 & 92.7 & 91.7 & 91.7 & 90.5 & 90.6 & 87.6 & 88.8 & 87.5 & 85.1 & 86.1 & 85.7 & 88.7 & 88.4 & 89.2 & 79.1 & 78.5 & 79.3 \\ 
    MNLI-m & 90.2 & 88.3 & 89.8 & 86.5 & 87.5 & 85.3 & 80.4 & 75.5 & 76.6 & 74.6 & 69.3 & 70.6 & 68.6 & 77.8 & 77.9 & 79.8 & 61.3 & 62.6 & 60.9 \\ 
    MNLI-mm & 90.2 & 88.5 & 89.5 & 86.2 & 87.8 & 85.0 & 81.1 & 77.1 & 77.4 & 75.5 & 70.6 & 71.2 & 70.8 & 79.7 & 78.9 & 80.8 & 62.6 & 62.3 & 62.3 \\ 
    CB & 90.5 & 80.6 & 90.5 & 88.0 & 69.9 & 84.6 & 78.7 & 63.7 & 77.3 & 80.5 & 63.5 & 63.6 & 77.3 & 83.5 & 83.2 & 84.8 & 63.8 & 60.5 & 57.7 \\ 
    ANLI-R1 & 73.8 & 75.9 & 66.6 & 56.1 & 60.1 & 58.5 & 52.6 & 47.7 & 48.1 & 46.6 & 45.6 & 48.6 & 47.4 & 49.5 & 47.2 & 51.1 & 37.8 & 39.8 & 39.7 \\ 
    ANLI-R2 & 48.9 & 57.7 & 44.6 & 41.5 & 41.5 & 42.9 & 44.5 & 38.8 & 40.9 & 39.1 & 39.7 & 42.0 & 40.5 & 40.4 & 40.2 & 41.3 & 36.1 & 36.6 & 36.4 \\ 
    ANLI-R3 & 44.4 & 54.4 & 41.3 & 40.8 & 42.0 & 43.8 & 42.1 & 38.3 & 40.8 & 40.7 & 38.0 & 38.2 & 38.3 & 40.7 & 41.0 & 41.7 & 31.8 & 34.2 & 33.5 \\ 
    \midrule
    COPA & 79.1 & 96.0 & 86.0 & 72.0 & 62.0 & 80.0 & 68.0 & 66.0 & 74.0 & 70.0 & 68.0 & 58.0 & 70.0 & 74.0 & 72.0 & 72.0 & 58.0 & 54.0 & 52.0 \\ 
    WSC & 89.0 & 78.8 & 78.8 & 76.9 & 61.5 & 65.4 & 57.7 & 61.5 & 61.5 & 59.6 & 55.8 & 59.6 & 50.0 & 59.6 & 55.8 & 61.5 & 59.6 & 51.9 & 67.3 \\ 
    CommonsenseQA & 72.1 & 80.5 & 74.6 & 58.5 & 23.8 & 60.8 & 57.4 & 33.8 & 36.1 & 32.5 & 26.1 & 26.1 & 23.3 & 41.3 & 38.7 & 43.0 & 21.6 & 22.5 & 25.1 \\ 
    MC-TACO & 44.0 & 55.9 & 55.9 & 48.6 & 47.7 & 43.2 & 38.7 & 37.8 & 40.5 & 37.8 & 35.1 & 34.2 & 28.8 & 37.8 & 41.4 & 40.5 & 20.7 & 27.0 & 24.3 \\ 
    SocialIQA & 78.5 & 79.4 & 79.9 & 70.5 & 38.9 & 66.1 & 59.9 & 54.4 & 55.2 & 56.0 & 51.8 & 49.9 & 50.8 & 58.5 & 56.9 & 55.6 & 43.5 & 46.1 & 44.1 \\ 
    WiC & 70.5 & 74.3 & 71.5 & 71.2 & 71.5 & 69.6 & 68.7 & 63.3 & 64.9 & 68.0 & 61.4 & 63.3 & 62.4 & 65.8 & 66.5 & 68.3 & 60.2 & 59.2 & 59.9 \\ 
    Abductive NLI & 83.9 & 83.8 & 85.0 & 72.2 & 76.4 & 66.4 & 62.4 & 56.9 & 57.4 & 56.7 & 55.0 & 54.8 & 55.0 & 57.7 & 56.1 & 60.3 & 52.9 & 53.3 & 53.9 \\ 
    PiQA & 77.1 & 80.6 & 77.6 & 68.2 & 53.5 & 65.7 & 59.7 & 60.7 & 61.7 & 61.5 & 60.6 & 60.4 & 60.0 & 62.9 & 60.3 & 62.2 & 55.0 & 55.9 & 55.9 \\ 
    Winogrande & 79.3 & 85.6 & 77.7 & 64.6 & 52.6 & 52.9 & 52.0 & 50.9 & 54.5 & 52.6 & 51.2 & 51.8 & 51.2 & 53.7 & 55.6 & 52.8 & 47.6 & 50.9 & 48.7 \\ 
    \midrule
    ARC-C & -- & 47.5 & 37.5 & 31.8 & 38.1 & 39.8 & 31.8 & 31.8 & 30.4 & 29.8 & 30.8 & 30.4 & 30.4 & 29.4 & 30.8 & 32.8 & 27.8 & 27.1 & 29.8 \\ 
    ARC-E & 66.0 & 69.3 & 62.5 & 53.2 & 40.9 & 61.4 & 56.8 & 39.1 & 41.6 & 40.0 & 37.0 & 36.0 & 34.0 & 41.8 & 43.0 & 46.5 & 31.8 & 31.9 & 29.8 \\ 
    ARCT & 70.1 & 83.5 & 86.7 & 76.3 & 76.6 & 74.4 & 72.2 & 66.5 & 68.0 & 66.5 & 64.6 & 66.1 & 66.5 & 70.3 & 68.7 & 70.3 & 66.1 & 64.6 & 64.6 \\ 
    MCScript & 90.0 & 96.0 & 92.8 & 85.0 & 90.1 & 84.1 & 80.7 & 74.1 & 74.7 & 73.5 & 68.9 & 70.5 & 69.7 & 73.7 & 76.2 & 77.8 & 60.6 & 61.7 & 60.1 \\ 
    BoolQ & 87.1 & 87.3 & 85.7 & 82.2 & 84.2 & 76.1 & 75.7 & 74.3 & 73.5 & 72.5 & 69.5 & 72.7 & 71.9 & 73.6 & 74.3 & 74.1 & 67.3 & 68.6 & 66.9 \\ 
    CosmosQA & 81.9 & 86.5 & 79.4 & 66.8 & 71.7 & 64.9 & 56.0 & 54.7 & 53.4 & 51.5 & 44.0 & 46.8 & 48.9 & 55.9 & 54.3 & 54.0 & 39.9 & 42.0 & 40.6 \\ 
    HellaSwag & 85.2 & 89.8 & 84.1 & 61.6 & 74.5 & 43.9 & 37.2 & 36.9 & 35.4 & 34.7 & 33.8 & 33.6 & 33.0 & 37.5 & 36.4 & 36.5 & 29.9 & 30.0 & 29.4 \\ 
    Mutual & 71.3 & 89.8 & 87.8 & 73.3 & 26.9 & 72.5 & 65.5 & 56.2 & 58.7 & 57.8 & 50.8 & 52.8 & 53.3 & 63.9 & 59.1 & 62.5 & 43.3 & 41.3 & 42.4 \\ 
    Mutual+ & 62.6 & 82.8 & 77.9 & 63.9 & 65.5 & 65.9 & 54.2 & 51.9 & 49.7 & 50.6 & 47.2 & 43.8 & 45.8 & 54.9 & 53.5 & 53.5 & 35.9 & 40.4 & 40.2 \\ 
    QuAIL & 47.9 & 78.0 & 73.3 & 67.0 & 63.9 & 52.9 & 54.4 & 52.2 & 53.6 & 53.6 & 47.8 & 50.0 & 48.5 & 54.2 & 54.1 & 57.3 & 40.7 & 38.7 & 41.8 \\
    \midrule
    QAMR & 79.1 & 79.6 & 79.6 & 77.6 & 79.3 & 76.4 & 73.5 & 70.1 & 71.8 & 70.4 & 63.4 & 64.4 & 62.6 & 73.7 & 73.1 & 74.6 & 24.0 & 28.7 & 27.8 \\ 
    NewsQA & -- & 59.6 & 57.8 & 53.1 & 56.2 & 53.5 & 48.0 & 36.2 & 38.5 & 34.7 & 29.1 & 29.6 & 27.5 & 40.4 & 42.4 & 45.9 & 6.5 & 8.9 & 9.7 \\
    SQuAD2.0 & 86.8 & 89.9 & 91.5 & 88.6 & 87.0 & 88.6 & 86.3 & 82.0 & 82.9 & 80.7 & 75.8 & 77.2 & 75.8 & 82.7 & 83.5 & 84.8 & 57.1 & 58.2 & 58.2 \\ 
    MRQA-NQ & 57.4 & 71.5 & 69.9 & 65.0 & 66.9 & 67.3 & 66.1 & 59.0 & 58.9 & 57.5 & 52.9 & 53.0 & 52.8 & 62.1 & 60.6 & 61.5 & 41.0 & 43.1 & 42.4 \\ 
    Quoref & 74.9 & 83.7 & 78.7 & 66.7 & 76.3 & 69.8 & 62.9 & 48.0 & 46.7 & 43.9 & 34.9 & 35.6 & 34.5 & 51.0 & 50.1 & 56.0 & 14.6 & 21.6 & 16.8 \\
    \bottomrule
    \end{tabular}
}
\caption{Results of all our models on each validation set. \textbf{RL}: RoBERTa$_{\rm{Large}}$, \textbf{RB}: RoBERTa$_{\rm{Base}}$, \textbf{BL}:BERT$_{\rm{Large}}$, \textbf{BB}:BERT$_{\rm{Base}}$, \textbf{RMS}: RoBERTa-Med-Small. \textbf{Best} denotes best known performance on the original dataset's validation set.}
\label{tab:result_all_tasks}
\end{sidewaystable*}

\end{document}